\documentclass{article}
\usepackage{ijcai24}

\usepackage{times}
\usepackage{soul}
\usepackage{url}
\usepackage[hidelinks]{hyperref}
\usepackage[utf8]{inputenc}
\usepackage[small]{caption}
\usepackage{graphicx}
\usepackage{amsmath}
\usepackage{amsthm}
\usepackage{booktabs}
\usepackage{algorithm}
\usepackage[switch]{lineno}

\usepackage{balance}

\usepackage{amsmath,amssymb,amsfonts}
\usepackage{graphicx}
\usepackage{textcomp}
\usepackage{xcolor}
\def\BibTeX{{\rm B\kern-.05em{\sc i\kern-.025em b}\kern-.08em
    T\kern-.1667em\lower.7ex\hbox{E}\kern-.125emX}}

\usepackage{caption}
\usepackage{multirow}
\usepackage{makecell}
\usepackage{subfigure}
\usepackage{bm}
\usepackage{booktabs}
\usepackage{pifont}%

\usepackage{algorithm,algpseudocode}
\usepackage{enumerate}
\usepackage{enumitem}
\usepackage{hyperref} 
\usepackage[nameinlink]{cleveref}
\usepackage{bbding}
\usepackage{url}
\usepackage{soul}
\newtheorem{remark}{\noindent \textbf{Remark}}

\usepackage{graphicx}
\usepackage{subcaption}

\newtheorem{defn}{\hspace{-1mm} \textsc{Definition}}

\def\eqref#1{equation~\ref{#1}}

\newcommand{\ours}{MSPT}

\title{Continual Multimodal Knowledge Graph Construction}

\author{
Xiang Chen$^{1,3}$\and
Jingtian Zhang$^{2,3}$\and
Xiaohan Wang$^{2,3}$\and
Ningyu Zhang$^{2,3}$\footnotemark[1]\and
Tongtong Wu$^{4}$\and \\
Yuxiang Wang$^{5}$\and
Yongheng Wang$^{6}$\And 
Huajun Chen$^{1,3}$\thanks{~~Corresponding author.}
\\
\affiliations
$^1$College of Computer Science and Technology, Zhejiang University \\
$^2$School of Software Technology, Zhejiang University \\ 
$^3$ZJU-Ant Group Joint Research Center for Knowledge Graphs, Zhejiang University,\\ 
$^4$Monash University, \\
$^5$Hangzhou Dianzi University, \\
$^6$Zhejiang Lab 
\\
\emails
{\{xiang\_chen, zhangjintian, wangxh07, zhangningyu, huajunsir\}@zju.edu.cn}\\
{tongtong.wu@monash.edu}, {lsswyx@hdu.edu.cn}, {wangyh@zhejianglab.com}
}

\begin{document}

\maketitle

\begin{abstract}
Current Multimodal Knowledge Graph Construction (MKGC) models struggle with the real-world dynamism of continuously emerging entities and relations, often succumbing to catastrophic forgetting—loss of previously acquired knowledge.
This study introduces benchmarks aimed at fostering the development of the continual MKGC domain.
We further introduce \textbf{\ours} framework, designed to surmount the shortcomings of existing MKGC approaches during multimedia data processing. \ours{} harmonizes the retention of learned knowledge (stability) and the integration of new data (plasticity), outperforming current continual learning and multimodal methods. Our results confirm {\ours}'s superior performance in evolving knowledge environments, showcasing its capacity to navigate balance between stability and plasticity.

\end{abstract}

\section{Introduction}

The rise of multimodal data on social media platforms has sparked significant interest among knowledge graph and multimedia researchers in the domain of multimodal knowledge graphs~\cite{MMKG,chengliang_kg,MMKGR,mm/xuming23,tkde/LiangLZTWYDL24}. To address the limitations of relying on human-curated multimodal data and to systematically extract insights from vast multimedia repositories, the concept of Multimodal Knowledge Graph Construction (MKGC) has been proposed~\cite{zhang2023multimodal,keliang_mm}. MKGC leverages multimodal data as an additional information source to disambiguate polysemous terms and perform tasks like Multimodal Named Entity Recognition (MNER)\cite{coling22-flat} and Multimodal Relation Extraction (MRE)\cite{multimodal-re}. However, existing MKGC architectures~\cite{multimodal-re,mkgformer} primarily focus on ``static'' knowledge graphs, where entity categories and relations remain fixed throughout the learning process. These models lack adaptability, especially when confronted with new entity categories and relations.

\begin{figure}[!]
    \centering
\includegraphics[width=0.48\textwidth]{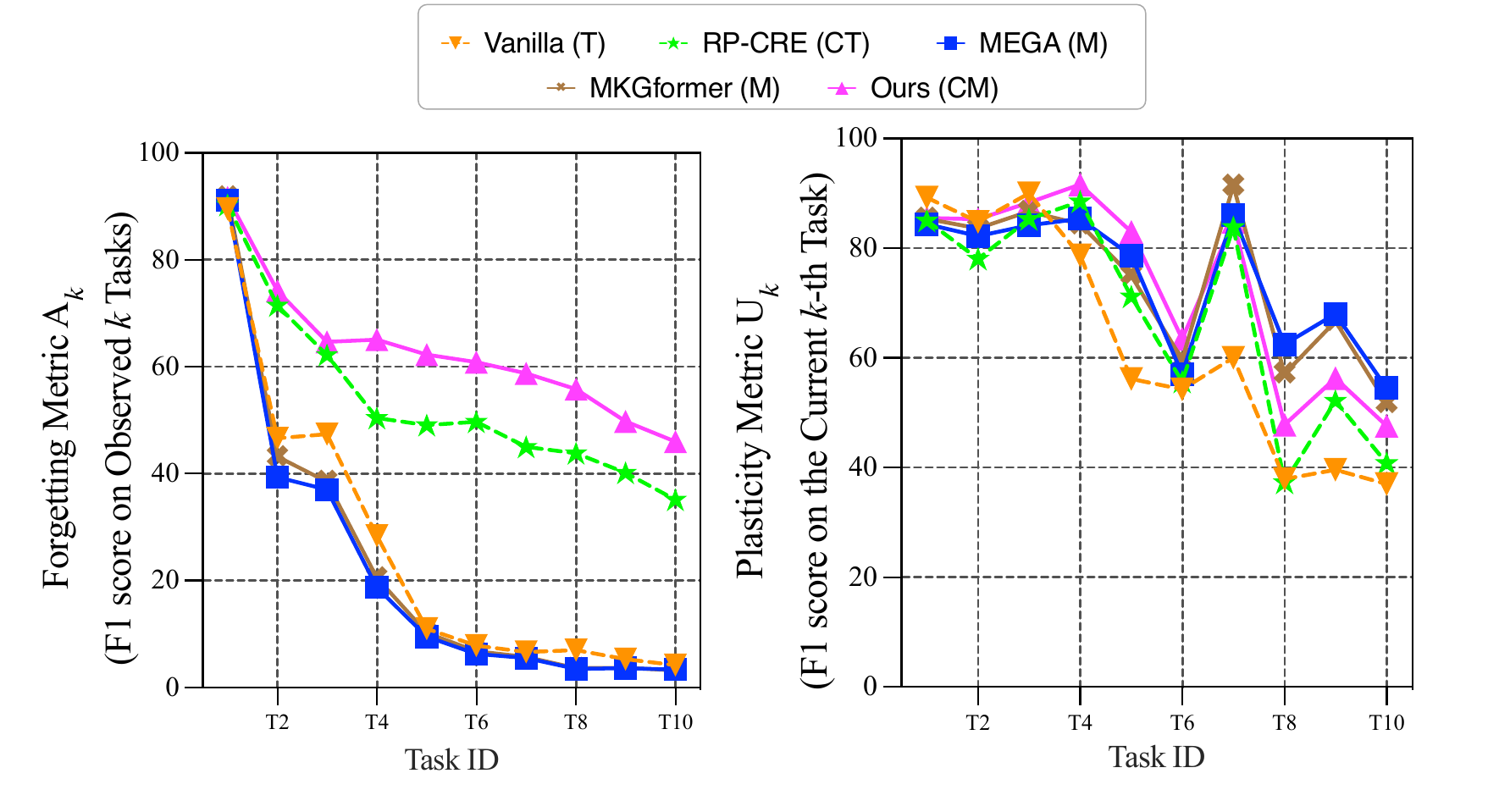}%
    \caption{
     Results on incremental MRE (IMRE) benchmark. We benchmark {\ours} against the Vanilla Training approach, multimodal KGC models such as MEGA and MKGformer, as well as the continual RE method RP-CRE. 
     }
    \label{fig:motivation}
\end{figure}

Addressing the dynamic nature of streaming data, replete with emerging entity categories and relations, the research community has developed the continuous Knowledge Graph Completion (KGC) methods~\cite{aaai21-continual,wang-etal-2022-shot,xia-etal-2022-learn}, seeking to balance the integration of new entity categories and relations (plasticity) with the preservation of established knowledge (stability). While current continual KGC strategies are largely text-centric, neglecting the demands of MKGC, the latter's capacity to handle multimodal data can provide richer insights than text-only models. Historical evaluations~\cite{chen-etal-2022-hvpnet,mm/xuming23} have demonstrated the superiority of MKGC in static KG settings.
However, the preliminary experimental results in Figure~\ref{fig:motivation} reveal significant hurdles when directly transferring MKGC models to a continual learning environment. Notably, MKGC models not only fall short of unimodal counterparts in previous tasks but also show limited effectiveness on current task test sets during continuous task training.

We posit that this observation may decline in MKGC models during continual learning may stem from disparate convergence rates among different modalities~\cite{cvpr20-imbalance-multimodal}, leading to two primary challenges:
\textit{Challenge 1: How to alleviate the
imbalanced learning dynamics across modalities to enhance the plasticity?} The differential learning dynamics in multimodal settings can hinder the adaptability of MKGC models to new entity categories and relations, especially when employing replay strategies. This imbalance may result in inferior representations, undermining the addition of new knowledge.
\textit{Challenge 2: How to reduce the
 forgetting in the process of multimodal interaction?} Continual MKGC models face the unique problem of varying forgetting rates across modalities, unlike their unimodal counterparts. These disparities can disproportionately affect secondary modalities, increasing the risk of forgetting and jeopardizing the performance of prior tasks.
Addressing these challenges necessitates the development of continual MKGC models that ensure uniform multimodal forgetting and robust modality integration to  manage the retention and acquisition of knowledge.


To overcome the highlighted challenges in continual MKGC, we introduce the Multimodal Stability-Plasticity Transformer (\textbf{\ours}), a novel framework that advances the stability-plasticity trade-off through strategic multimodal optimization. Our method is distinguished by two pivotal modules:
\textit{(1) Gradient Modulation for Balanced Learning:} We propose a gradient modulation technique to address the imbalanced learning dynamics across modalities, thereby preserving the model's ability to learn new information. By adaptively tuning gradients according to each modality's optimization contribution, our approach ensures nuanced representation development for both modalities, enhancing plasticity.
\textit{(2) Hand-in-Hand Multimodal Interaction with Attention Distillation:} Deviating from traditional cross-attention multimodal interaction, \ours{} calculates inter-modal self-query affinities against an external learnable key. This decoupling of fusion parameters allows for a more deliberate modulation of forgetting rates, promoting consistent knowledge retention. And the attention distillation is utilized to refine this process, leveraging the multimodal interaction outputs to preserve crucial attention patterns. 
The results of our thorough experiments demonstrate that \textbf{\ours} outperforms both traditional MKGC and continual unimodal KGC models in various class-incremental settings, showcasing its potential in the field of continual MKGC\footnote{Our data and code are available at \url{https://github.com/zjunlp/ContinueMKGC}}.








\section{Related Works}
\label{sec:related}

\subsection{Advancements in MKGC}


\paragraph{Multimodal Named Entity Recognition.}
Advancements in MNER have shifted from text-only approaches to also harnessing visual cues. Studies such as those by \cite{zhang2018adaptive,lu2018visual,moon2018multimodal,arshad2019aiding}  have introduced interactions between CNN-driven visual and RNN-based textual features. Others, UMT~\cite{yu-etal-2020-improving} and UMGF~\cite{zhang-UMGF}, have suggested utilizing fine-grained semantic correspondences with a combination of transformer and visual backbones, taking into account regional image features to represent objects. The ITA~\cite{ITA} model exploits self-attention to enrich text embeddings with image spatial context, showing superiority over text-centric models.

\paragraph{Multimodal Relation Extraction.}
Researchers have started exploring techniques to link entities mentioned in the textual content with corresponding objects depicted in associated images. 
Some examples include work done by \cite{9428274}, who presents an MRE dataset that can associate the textual entities and visual objects for enhancing relation extraction.
Then \cite{multimodal-re}  revises the  MRE dataset based on \cite{9428274} and utilizes scene graphs to align textual and visual representations.
\cite{FL-MSRE} also collects and labels four MRE datasets based on four famous works in China to address the scarcity of resources for multimodal social relations.


\subsection{Continual Knowledge Graph Construction}


Continual learning addresses catastrophic forgetting in the following strategies: consolidation-based methods~\cite{icml-synaptic,liangke_sigir23} that adjust parameter updates through regularization, dynamic architectures~\cite{Progressive} that evolve with data, and rehearsal-based methods~\cite{iclr-memory-adap,iclr-gem} using memory banks to preserve knowledge. The latter has exhibited superior performance in continual KGC~\cite{aaai21-continual}.
To address the challenge of continual RE, memory interaction methods~\cite{RP-CRE-ACL2021-refining} have been proposed to effectively utilize representative samples. Additionally, prototype methods~\cite{EMAR-han-etal-2020-continual,RP-CRE-ACL2021-refining} are increasingly employed to abstract relation information and mitigate overfitting.
In the context of continual NER, the ExtendNER method~\cite{aaai21-continual} tackles class-incremental learning by creating a unified NER classifier that encompasses all encountered classes over time. Moreover, approaches \cite{xia-etal-2022-learn,wang-etal-2022-shot} prevent forgetting of previous NER tasks by utilizing stored or generated data from earlier tasks during training.
However, previous studies have focused on continual KGC and have not been readily applicable to  MKGC due to the inherent challenges posed by multimodal data.

\section{Preliminaries}
\label{sec:perliminary}



\subsection{Delineation of MKGC Tasks}

\begin{defn}{{MNER.}}
 This subtask emphasizes the extraction of named entities from textual content and its associated images. Given a token sequence, denoted as \(x^t = [w_1, \ldots, w_m]\), and its affiliated image patch sequence \(x^v\), the principal goal of continual MNER is to consistently model the sequence tags' distribution, expressed as \(p(y | (x^t, x^v))\). Within this context, for task \(T_k\), the label sequence \(y\) is defined as \(y = [y_1, \ldots, y_m]\) and integrates emergent entity types from the entity category set 
 \(\mathcal{E}_k\).
\end{defn}

\begin{defn}{{MRE.}}
 This subtask focuses on extracting relationships between designated entity pairs from token sequences. For a given task \(T_k\), and provided with a token sequence \(x^t\) and its corresponding image patch sequence \(x^v\), the goal is to infer the relationship of a specific entity pair, \((e_h, e_t)\),  derived from \(x^t\). A key challenge lies in computing the probability distribution over possible relations \(r\) from the set \(\mathcal{R}_k\), expressed as \(p(r | (x^t, x^v, e_h, e_t))\). This is made more complex by the potential addition of novel relations to \(\mathcal{R}_k\).
\end{defn}

\subsection{Class-Incremental Continual Learning}
We define a class-incremental continual learning scenario as a series of $K$ separate tasks, each with its schema classes and MKGC corpus. Formally, the tasks are denoted as:
\begin{equation}
\footnotesize
\mathcal{T} = [(\mathcal{S}^1, C^1), (\mathcal{S}^2, C^2), ..., (\mathcal{S}^K, C^K)].
\end{equation}
The $k$-th task $T_k$ includes a distinct set of entity types $\mathcal{E}_k$ and relations $\mathcal{R}_k$, along with an MKGC corpus $C^k$ which is divided into training, validation, and testing subsets $\mathcal{D}_k$, $\mathcal{V}_k$, and $\mathcal{Q}_k$, respectively.
Each training instance in $\mathcal{D}_k$ consists of a textual input $x^t$, a sequence of image patches $x^v$—utilizing ViT encoding—and a corresponding label $y$, which is either an entity from $\mathcal{E}_k$ or a relation from $\mathcal{R}_k$.
Learners are restricted to use only the data from $\mathcal{D}_k$ during the training phase of task $T_k$, and to ensure non-overlapping classes between tasks, we enforce $\mathcal{E}_i \cap \mathcal{E}_j = \emptyset$ and $\mathcal{R}_i \cap \mathcal{R}_j = \emptyset$ for $i \neq j$. This setup follows the convention of several benchmark methodologies~\cite{cil_survey}.
In our class-incremental MKGC setting, after training completes on $\mathcal{D}k$, the model undergoes evaluation across an aggregated test set $\cup{i=1}^k\mathcal{Q}_i$, which includes all class categories up to the current task. This differs from task-incremental learning, where evaluation is confined to the specific task $\mathcal{S}^{k}$.
The evaluation metrics are introduced as follows:

\begin{defn}
Forgetting Metric (\(\mathbf{A}_k\)): ** 
 Measures the \(F1\)  score on aggregate test sets \(\bigcup_{i=1}^k \mathcal{Q}_i\) for tasks \(\{T_i\}_{i=1}^{k}\) post-training on \(T_k\). It indicates the model's ability to prevent catastrophic forgetting, especially in sequential data with new entity categories and relations.
\end{defn}

\begin{defn}
Plasticity Metric (\(\mathbf{U}_k\)): 
** Defined by the \(F1\)  score on the current task \(T_k\), showcasing the model's capacity to learn new tasks while retaining existing knowledge, a critical aspect of continual learning.
\end{defn}

\section{Methodology}
\label{sec:framework}

\subsection{Framework Overview}
\begin{figure*}[t!]
    \centering
            \subfigure[Hand-in-hand Multimodal Interaction with Attention Distillation.]{
    \includegraphics[width=0.6\textwidth]{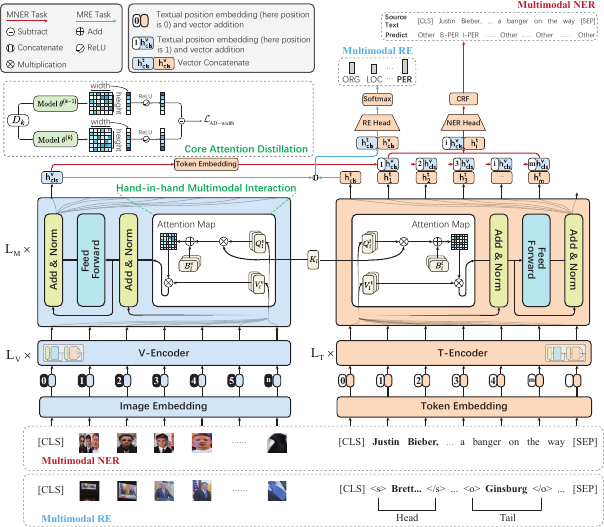}
    \label{fig:arc1}
    }
            \subfigure[Balanced Multimodal Learning Dynamics.]{
    \includegraphics[width=0.32\textwidth]{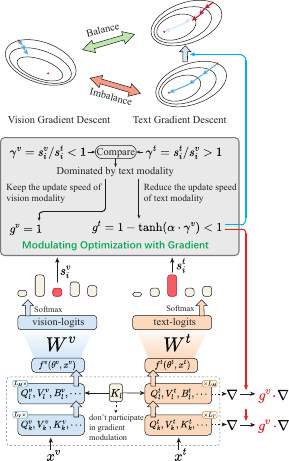}
    \label{fig:arc2}
    }
    \caption{Overview of our {\ours} framework. 
    \label{fig:arc}}
\end{figure*}

As illustrated in Figure~\ref{fig:arc}, 
our continual KGC framework adopts a dual-stream Transformer structure with the task-specific paradigm, including: 

\textbf{(1) Structure.}
 We incorporate a Visual Transformer (ViT)~\cite{vit} for visual data and BERT for textual data. Building on prior research~\cite{DBLP:conf/blackboxnlp/ClarkKLM19,lighter}, which indicates that manipulating the upper layers of language models (LMs) more effectively leverages knowledge for downstream tasks, our framework engages in multimodal interactions and attention distillation within the top three layers of the Transformers.

\textbf{(2) Task-specific paradigm.}  For the MRE task, we employ a task-specific approach by fusing the $\texttt{[CLS]}$ token representations from both ViT and BERT models. This integrated representation enables us to derive the probability distribution over the relation set $\mathcal{R}$  for the given task.
\begin{equation}
\footnotesize
p (r | ( x^t, x^v, e_h, e_t)) = \texttt{Softmax}({W}\cdot  [\bm{h}^t_{cls};\bm{h}^v_{cls}]),
\end{equation}
 where $\bm{h}^t \in \mathbb{R}^{m_t \times d_t}$ and $\bm{h}^v \in \mathbb{R}^{m_v \times d_v}$ represent the output sequence embeddings from BERT and ViT, respectively. In the context of MNER, for fair benchmarking against prior work, we employ a CRF function akin to that in the {\ours} framework. For the entity tag sequence $y = [y_1, \ldots, y_m]$, we enhance the BERT embeddings with $\bm{h}^v_{cls}$ and positional embeddings $\mathbb{E}_{\text{pos}}^t$ to capture visual information. The probability of a tag sequence $y$ within the predefined label set $Y$ is computed using the BIO tagging scheme that follows in~\cite{lample2016neural} as:
\begin{equation}
\footnotesize
p( y_i | ( x^t, x^v)) = \texttt{Softmax}({W} \cdot [\bm{h}_{i}^{t}; (\bm{h}^{v}_{cls}+\mathbb{E}_{\texttt{pos}_i}^{t})]).
\end{equation}

\subsection{Balanced Multimodal Learning Dynamics}
\label{sec:m-encoder}

{\textbf{Modulating Optimization with Gradient}.}
As elucidated in \S~Appendix-A, diverse convergence rates across modality-specific parameters can lead to imbalanced learning dynamics during continual learning, potentially hampering current task performance. To address this, we propose a gradient modulation strategy to fine-tune the optimization of visual and textual encoders, depicted in Figure~\ref{fig:arc2}. Building upon concepts, we adapt these to the $k$-th task using the \textit{Stochastic Gradient Descent} (SGD) algorithm:
\begin{equation}
\label{update_phi}
\footnotesize
\begin{aligned}
\theta_{n+1}^{v(k)} &=\theta_{n}^{v(k)}-\eta\nabla_{\theta^{v}} \mathcal{L}_{CE}(\theta_{n}^{v(k)})\\
&=\theta_{n}^{v(k)}-\eta\varphi(\theta_{n}^{v(k)})
\end{aligned}
\end{equation},
where $\varphi(\theta_{n}^{v(k)})=\frac{1}{N}\sum_{x\in  B_{n}\nabla_{\theta^{v}}} \ell (x;\theta_{n}^{v(k)})$ is an unbiased estimation of the full gradient,  $B_n$ represent a random mini-batch with $N$ samples at step $n$ of optimization, and $\nabla_{\theta^{v}} \ell (x;\theta_{n}^{v(k)})$ denotes the gradient w.r.t. batch $B_n$.

Drawing from \cite{Peng2022Balanced}, to counteract imbalanced multimodal learning dynamics, we introduce an adaptive gradient modulation mechanism for visual and textual modalities. This is based on quantifying their respective contributions to the learning goal via the contribution ratio $\gamma_n$:
\begin{equation}
\label{8}
\footnotesize
\begin{gathered}
s_{i}^{v}=  \sum_{y=1}^M 1_{y=y_{i}} \cdot  \text{softmax} (W^{v}_{n} \cdot f^{v}_{n}(\theta^{v},x_{i}^{v}))_{y}, \\
s_{i}^{t}=  \sum_{y=1}^M 1_{y=y_{i}} \cdot  \text{softmax} (W^{t}_{n} \cdot f^{t}_{n}(\theta^{t},x_{i}^{t}))_{y},
\end{gathered}
\end{equation}
\begin{equation}
\label{calcu_ratio}
\footnotesize
\gamma^{t}_{n}=\frac{\sum_{i \in B_{n}}  s_{i}^{t} } {\sum_{i \in B_{n}}  s^{v}_i}.
\end{equation}
To dynamically assess the contribution ratio $\gamma^{t}_{n}$ between textual and visual modalities, we introduce a modulation coefficient $g_n^t$ that adaptively regulates the gradient, defined as:
\begin{equation}
\label{gn}
\footnotesize
\begin{gathered}
g_n^t=\left\{\begin{array}{cl}1-\tanh \left(\alpha \cdot \gamma_n^t\right) & \gamma_n^t>1 \\ 1 & \text { otherwise}\end{array}\right.,
\end{gathered}
\end{equation}
\begin{equation}
\label{gt}
\footnotesize
\begin{gathered}
G^{t{(k)}}=\text{Avg}(g_n^{t(k)}),
\end{gathered}
\end{equation}
where $\alpha$ is a hyper-parameter that adjusts the influence of modulation.
, $G^{t{(k)}}$ is the averaged modulation coefficient of the model trained after the task $k$.
We further propose to balance the multimodal learning rhythm by integrating the coefficient $g_n^t$ into the SGD optimization process of task $k$ in iteration $n$ as follows:
\begin{equation}
\label{cx2}
\footnotesize
\theta_{n+1}^{v(k)}=\left\{
\begin{array}{cl} \theta_{n}^{v(k)}-\eta {g_n^{t(k)}} \varphi(\theta_{n}^{v(k)}) 
& k=1 \\
\theta_{n}^{v(k)}-\eta G^{t{(k-1)}} \varphi(\theta_{n}^{v(k)}) & k>1 \end{array}\right..
\end{equation}

\begin{remark}
Through gradient modulation in task 1 and using the average coefficient from the preceding $k-1$ tasks to influence training on the current $k$-th task, we ensure a smoother transition in balanced multimodal learning across tasks.
\end{remark}

\subsection{Hand-in-hand Multimodal Interaction via Attention Distillation}
Inspired by the concept of collaborative progress with the saying goes \textit{``hand in hand, no one is left behind''}, 
our method establishes a coherent framework for continual learning by integrating a dual-stream Transformer with attention distillation. As depicted in Figure~\ref{fig:arc1}, the model promotes uniform learning across various modalities, reducing unequal forgetting and enhancing multimodal resilience.

\begin{table*}[]
\centering
\small
\scalebox{0.75}{
\begin{tabular}{cccccccccccc}
\toprule
\multicolumn{12}{c}{\textbf{IMRE Benchmark}} \\ 
 \midrule
\multicolumn{1}{c|}{Model} &  \multicolumn{1}{c|}{Resource}
& T1 & T2 & T3 & T4 & T5 & T6 & T7 & T8 & T9 & T10 \\ 
    \midrule
 \multicolumn{1}{c|}{Vanilla} 
& \multicolumn{1}{c|}{Text}
&89.6    &46.6    &47.4    &28.4     &11.0   &7.8     &6.6    &7.0     &5.3     &4.2      \\
 \multicolumn{1}{c|}{EWC~\cite{ewc}} 
& \multicolumn{1}{c|}{Text}  
& 89.2    & 38.4   & 37.8   & 37.0   & 10.4    & 7.1   & 5.8   & 2.2   & 2.9   & 3.0  
\\
 \multicolumn{1}{c|}{EMR ~\cite{wang-etal-2019-sentence}} 
& \multicolumn{1}{c|}{Text} 
& 84.8   & 56.3   &  42.4  & 30.6   & 19.3   & 19.8   & 13.0   & 9.8   & 9.6   &   10.9
\\
 \multicolumn{1}{c|}{EMAR-BERT~\cite{EMAR-han-etal-2020-continual}} 
& \multicolumn{1}{c|}{Text} 
&90.3    & 57.0   &61.1    &49.1    &48.4    &45.5    &37.6    &31.0    &30.1    & 25.8  
\\
 \multicolumn{1}{c|}{RP-CRE~\cite{RP-CRE-ACL2021-refining}} 
& \multicolumn{1}{c|}{Text} 
& 90.3   &71.5    &62.2    &50.4   &49.1   &49.7    &45.0    &43.8    & 40.2   &35.1   
\\
    \midrule

 \multicolumn{1}{c|}{UMT~\cite{yu-etal-2020-improving} }
& \multicolumn{1}{c|}{Text+Image} 

&  90.2  & 38.2   &  37.2  &  18.3  & 10.3   & 6.5   & 5.5   & 3.2   &  3.3  & 2.8 
 \\
 \multicolumn{1}{c|}{UMGF~\cite{zhang-UMGF} }
& \multicolumn{1}{c|}{Text+Image}

& 90.7   &  41.5  & 38.3   &  18.8  & 10.7   & 6.0   & 5.2   & 3.3   & 3.5   & 3.0 
\\
 \multicolumn{1}{c|}{MEGA~\cite{multimodal-re} }
& \multicolumn{1}{c|}{Text+Image} 

&91.1   &39.3    &38.0    &19.7    &9.9  &6.3   &5.9    &3.5    & 3.6   &3.3    
\\ 
 \multicolumn{1}{c|}{MKGformer~\cite{mkgformer}}
& \multicolumn{1}{c|}{Text+Image} 

&\textbf{91.8}    &43.2    &38.5    &20.5    &11.2    &6.7     &5.7    &3.6    &3.7    &3.2  
\\ 
 \multicolumn{1}{c|}{M-[EWC]}
& \multicolumn{1}{c|}{Text+Image} 

&91.8    &46.3    &40.0    &31.8    &11.8    &7.4    &6.8    &4.5    &5.8    &3.9
\\
 \multicolumn{1}{c|}{M-[R-Replay]}
& \multicolumn{1}{c|}{Text+Image} 

& 91.8   & 60.2   & 61.4   & 50.5   & 43.1    &40.0   &34.5    &31.7    & 28.4   &30.8
\\ 
\multicolumn{1}{c|}{M-[R-Replay \& EWC]}
& \multicolumn{1}{c|}{Text+Image} 

& 91.8   & 62.3   & 57.6   & 49.5   & 44.5    &38.7    &31.9    &34.3    & 31.3   &28.0
\\ 
\multicolumn{1}{c|}{Joint Training}
& \multicolumn{1}{c|}{Text+Image} 
& 91.8 & 82.0 & 82.6 & 79.2 & 77.2 & 75.8 & 71.4 & 72.6 & 67.8 & 67.0 \\
 \midrule

 \multicolumn{1}{c|}{{\ours}}
& \multicolumn{1}{c|}{Text+Image}

& 91.7   & \textbf{76.2}   & \textbf{64.7}   &  \textbf{65.1}  &  \textbf{62.3}  & \textbf{60.9}   & \textbf{58.8} & \textbf{55.8}   & \textbf{49.8}   &  \textbf{46.0}
\\
\bottomrule
\end{tabular}
}
\caption{ Forgetting Metric  $\mathbf{A}_k$ (F1 score (\%) ) on  IMRE benchmark. The best performance results other than the upper bound model (Joint Training) are bolded. We explain uniformly that ``R-Replay'' refers to replay with random sampling. ``M-[]'' denotes the continual learning strategies applied to the multimodal method (here is MKGformer).}
\label{tab:exp-mnre}
\end{table*}

\subsubsection{\textbf{Hand-in-hand Multimodal Interaction}}
The self-attention mechanism (SAM)~\cite{transformer}, central to Transformer-based architectures, derives attention maps through self-key and self-query similarity calculations. Our proposed multimodal interaction approach introduces a unique attention generation process using shared learnable keys ($K_W$) and corresponding self-queries to enhance knowledge consolidation and retention. This method aims to counteract catastrophic forgetting by embedding previous task knowledge into the attention framework. It also promotes a tighter integration between visual and textual encoders, minimizing fusion bias and inconsistency associated with forgetting. Additionally, by regulating updates to $K_W$, our strategy preserves knowledge from earlier tasks, safeguarding against information degradation during new task acquisition.

Applying linear transformations to the input tensors $X_v$ and $X_t$, we obtain the visual self-query $Q^{X_v}=W^{q_v} X_v$ and self-value $V^{X_v}=W^{v_v} X_v$, alongside the textual self-query $Q^{X_t}=W^{q_t} X_t$ and self-value $V^{X_t}=W^{v_t} X_t$ using the visual and textual encoders' parameters $W^{q_v}, W^{v_v}, W^{q_t}, W^{v_t}$, respectively. We introduce a shared external key $K^s$ that supersedes the original self-key, generating updated attention maps for both modalities.
For the $k$-th task, utilizing a ViT and BERT model, we denote the prescaled attention matrix at the $l$-th layer as $A_l^{(k)}$ and the resulting SAM output as $Z_l^{(k)}$, prior to softmax activation. Note that details on multi-head attention and normalization are omitted for conciseness.
\begin{equation}
\footnotesize
\begin{aligned}
{\mathrm{A}_{l}^v}^{(k)}
=\frac{Q_l^{X_v}\left(K^s_l\right)^{\top}+{B_l^v}}{\sqrt{d_v / H}}, 
{\mathrm{A}_{l}^t}^{(k)}
=\frac{Q_l^{X_t}\left(K^s_l\right)^{\top}+{B_l^t}}{\sqrt{d_t / H}}, 
\end{aligned}
\end{equation}
\begin{equation}
\footnotesize
\begin{aligned}
{Z^{v}_{l}}^{(k)}  &=\texttt{Softmax}\left(\mathrm{A}_l^{v(k)}\right) V_l^{X_v}, \\
{Z^{t}_{l}}^{(k)} &=\texttt{Softmax}\left(\mathrm{A}_l^{t(k)}\right) V_l^{X_t}, 
\quad l=L-2, \ldots, L,
\end{aligned}
\end{equation}
where $L$ denotes the encoder's total number of layers, and $B_l^v$ and $B_l^t$ serve as bias terms for the vision and text attention maps, respectively. Nnote that the external key $K^s_l$ is not constrained by the current feature input, allowing for end-to-end optimization and the integration of prior knowledge.

\subsubsection{\textbf{Core Attention Distillation}}
\label{sec:transformer}

 Accordingly, we introduce a method for refining the attention matrices within the dual-stream Transformer's interaction module. This involves stabilizing attention maps through a distillation function that leverages learnable shared keys $K^s$ to mitigate forgetting. Considering the attention maps on the\textit{visual side}  across consecutive steps $k$ and $(k-1)$, we quantify the distillation loss in the width dimension as:
\begin{equation}
\footnotesize
\begin{aligned}
\mathcal{L}_{\text{AD-width }}^v\left( {\mathrm{A}_{l}^v}^{(k-1)}, {\mathrm{A}_{l}^v}^{(k)}  \right)=\sum_{h=1}^{H} {\delta}_{W}\left( {\mathrm{A}_{l}^v}^{(k-1)}, {\mathrm{A}_{l}^v}^{(k)}  \right),
\end{aligned}
\end{equation}
where $H$ and $W$ represent the height and weight of the attention maps. The total distance between attention maps $a$ and $b$ along the $h$ or $w$ dimension is represented by ${\delta}(a,b)$.

Our proposed attention distillation framework incorporates a crucial asymmetric distance function $\delta$, diverging from typical continual learning approaches that use symmetric Euclidean distance for model outputs comparison across tasks $k$ and $(k-1)$\cite{wang-etal-2022-shot,DBLP:conf/eccv/DouillardCORV20}. Symmetric distances tend to equally penalize shifts in attention from both new and old tasks, potentially impeding learning by increasing the loss when attention to previous tasks is maintained. Although preserving past tasks' knowledge mitigates forgetting, over-penalization can inadvertently suppress newly acquired insights, creating a tension between preserving past knowledge and embracing new information. To address this, we suggest $\delta$, an asymmetric distance measure that conserves prior knowledge while sustaining the model's adaptability, aligning with findings in computer vision\cite{AttentionDistill}. 
The modified function, ${\delta}_W$, is specifically crafted to balance the trade-off between plasticity and forgetting, illustrated as follows:
\begin{equation}
\footnotesize
\begin{aligned}
\delta_W\left({\mathbf{A}_{l}^v}^{(k-1)}, {\mathbf{A}_{l}^v}^{(k)}\right)=\left\|\mathcal{F}_{\text {asym }}\left(\sum_{h=1}^{H} {\mathbf{A}^{v(k-1)}_{l_{w, h}}}-\sum_{h=1}^{H} {\mathbf{A}^{v(k)}_{l_{w, h}}}\right)\right\|.
\end{aligned}
\end{equation}
We employ $\mathcal{F}_\text{asym}$, an asymmetric distance function, with ReLU~\cite{DBLP:conf/icml/NairH10} integrated as $\mathcal{F}_{asym}$ in subsequent experiments. The attention distillation loss is:
\begin{equation}
\label{loss:distill}
\begin{aligned}
\mathcal{L}_{\text{AD}} &= \mathcal{L}_{\text{AD-width }}^v +\mathcal{L}_{\text{AD-width}}^t.
\end{aligned}
\end{equation}

\begin{table*}[htbp!]
  \centering
    \scalebox{0.65}{
    \begin{tabular}{c|c|llll|llll}
   \toprule
        \multirow{2}{*}{Model} &  \multirow{2}{*}{Resource} 
          & \multicolumn{8}{c}{\textbf{IMNER Benchmark}}
          \\
          & 
          & \multicolumn{1}{c}{PER} $\rightarrow$
          & \multicolumn{1}{c}{ORG} $\rightarrow$
          & \multicolumn{1}{c}{LOC} $\rightarrow$
          & \multicolumn{1}{c|}{MISC}   
         & \multicolumn{1}{c}{PER} 
         $\rightarrow$
         & \multicolumn{1}{c}{LOC} 
         $\rightarrow$          
           & \multicolumn{1}{c}{ORG} 
         $\rightarrow$
          & \multicolumn{1}{c}{MISC}       \\
    \midrule

 \multicolumn{1}{c|}{Vanilla} 
 & \multicolumn{1}{c|}{Text}  
& 74.2  &38.9    &20.8    &12.6     &74.2    & 26.9    &  35.0  &   12.8       \\

 \multicolumn{1}{c|}{EWC~\cite{ewc}} 
& \multicolumn{1}{c|}{Text}  
& 76.2   & 40.5   & 20.7   & 14.4   & 76.3    & 28.9    & 36.9    & 13.4   
\\
 \multicolumn{1}{c|}{EMR~\cite{wang-etal-2019-sentence}} 
& \multicolumn{1}{c|}{Text}  
&  72.5  & 45.8   & 35.3   & 20.5    &   72.5 & 43.2   &  48.7  & 22.5   

\\
 \multicolumn{1}{c|}{EMAR-BERT~\cite{EMAR-han-etal-2020-continual}} 
& \multicolumn{1}{c|}{Text}  
&  73.2 & 48.5   & 42.5   & 28.7   & 73.2    & 45.5   & 50.3   &    30.8

\\
 \multicolumn{1}{c|}{ExtendNER~\cite{aaai21-continual} } 
& \multicolumn{1}{c|}{Text} 
& 50.7   &53.3    & 47.8   & 41.4   &55.6    & 52.9   &57.9    &47.9  
\\
    \midrule

 \multicolumn{1}{c|}{UMT~\cite{yu-etal-2020-improving} }
& \multicolumn{1}{c|}{Text+Image} 
&71.2    &37.3    & 19.1   &11.6    &71.2    &24.9    &33.6    & 11.6   
 \\
 \multicolumn{1}{c|}{UMGF~\cite{zhang-UMGF} }
& \multicolumn{1}{c|}{Text+Image}
& 73.0   & 39.1   &18.6    &10.9    & 73.1   &26.9    & 34.3   & 12.6    
 \\
 \multicolumn{1}{c|}{MEGA~\cite{multimodal-re} }
& \multicolumn{1}{c|}{Text+Image} 
& 72.5  & 38.7   &  18.9  & 11.2   & 72.3 & 24.6  & 33.8  & 11.5     
\\ 
 \multicolumn{1}{c|}{MKGformer~\cite{mkgformer}}
& \multicolumn{1}{c|}{Text+Image} 
& 77.6   &  38.4  &19.0    &11.8   &74.5   & 25.2    &34.0    & 11.3   
\\ 
 \multicolumn{1}{c|}{M-[EWC]}
& \multicolumn{1}{c|}{Text+Image} 

&74.5    &39.9    & 20.9   & 13.9   & 74.5    & 28.3   & 35.1   & 13.4    \\
 \multicolumn{1}{c|}{M-[R-Replay]}
& \multicolumn{1}{c|}{Text+Image} 

& 74.5   & 37.6    & 22.3   & 15.7   & 74.5   & 28.6   & 42.1   & 19.5   
\\
\multicolumn{1}{c|}{M-[R-Replay \& EWC]}
& \multicolumn{1}{c|}{Text+Image} 

& 74.5   & 41.5    & 24.6   & 21.0   & 74.5   & 28.6   & 40.7   & 17.0   
\\
\multicolumn{1}{c|}{Joint Training}
& \multicolumn{1}{c|}{Text+Image}
& 77.6 & 71.2 & 73.7 & 69.1 & 79.8 & 76.2 & 71.7 & 69.1
\\
 \midrule

\multicolumn{1}{c|}{\ours}
& \multicolumn{1}{c|}{Text+Image}
& \textbf{79.8}    & \textbf{66.4}  & \textbf{63.2}   & \textbf{48.2}   & \textbf{77.8}   & \textbf{62.6}   & \textbf{64.8}   & \textbf{62.3}
\\
\bottomrule
\end{tabular}
} 
\caption{Forgetting Metric  $\mathbf{A}_k$ (F1 score (\%) ) on  IMNER benchmark with two different order perturbations.}
\label{tab:exp_mner}
\end{table*}

\begin{remark}
This setup permits the development of new attention patterns during the $k$-th task without penalties, while attention absent in the current but present in the $(k-1)$-th task is penalized, promoting targeted knowledge retention.
\end{remark}

\subsection{Training Objective}
Our model leverages a cross-entropy loss ($\mathcal{L}{\text{CE}}$) to effectively recognize entities and relations, while an attention distillation loss ($\mathcal{L}{\text{AD}}$) mitigates the issue of catastrophic forgetting. We formulate the combined loss function as:

 \begin{equation}
\label{loss:all}
\footnotesize
\begin{aligned}
\mathcal{L}_{\text{all}} &=\lambda  \mathcal{L}_{\text{AD}} +\mathcal{L}_{\text{CE}}.
\end{aligned}
\end{equation}
Here, $\lambda$ serves as the weighting factor for the attention distillation loss. Additionally, we adopt the rehearsal strategy from PR-CRE to retain a concise memory set—merely six examples per task—for continual learning alignment, and optimizing memory footprint. The training protocol, inclusive of the rehearsal mechanism, is delineated in \S Appendix-B.

\section{Experiments}
\label{sec:experiment}
\subsection{ Incremental MKGC Benchmarks}

\paragraph{\textbf{IMNER Benchmark.}}
We utilize the established Twitter-2017 MNER dataset, which consists of multimodal tweets from 2016-2017, containing examples with multiple entity categories. To simulate more realistic learning conditions and reduce labeling ambiguity, we transition to a class-incremental framework, modifying the dataset such that each entity category is exclusive to a single task.

\paragraph{\textbf{IMRE Benchmark.}}
For our IMRE benchmark, we partition the dataset into 10 subsets for 10 distinct tasks. The original benchmark imposes two constraints that are at odds with the principles of lifelong learning: (1) clustering semantically related relations, and (2) excluding the ``N/A'' (not applicable) class. To rectify this, every task incorporates the ``N/A'' class, and relations are randomly sampled without bias, enhancing the benchmark's diversity and adherence to real-world lifelong learning conditions.

\begin{table}[t!] 
\scalebox{0.64}{
\begin{tabular}{cccc|cccc|c} 
\toprule
 \multicolumn{4}{c}{ Module }  & \multicolumn{5}{c}{ IMNER } \\
MI & AD & GM & MM  & {PER} $\rightarrow$ & LOC  $\rightarrow$ & ORG  $\rightarrow$ & MISC & AVG   \\
\hline & \Checkmark & \Checkmark & \Checkmark & 76.9 &\textbf{71.1} & 54.6  & 41.9 & 61.1\\
\Checkmark & & \Checkmark & \Checkmark & 77.8 & 42.0 & 64.7 & 41.2 &56.4 \\
\Checkmark & \Checkmark & & \Checkmark & 69.1 & 35.8 & 54.8 & 18.6 & 44.6 \\
\Checkmark & \Checkmark & \Checkmark & & 77.8 & 27.5 & 36.1 & 11.9 & 38.3  \\
\Checkmark & \Checkmark & \Checkmark & \Checkmark & \textbf{79.5} & 62.6 & \textbf{64.8} & \textbf{62.3} & \textbf{66.8} \\
\bottomrule
\end{tabular}
}
\caption{\label{ablation}
Ablation Study. ``MI'': Multimodal Interaction;
``AD'': Attention Distillation; 
``GM'': Gradient Modulation; ``MM'': Memory.}
\end{table}

\subsection{Compared Baselines}
We benchmark our {\ours} against SOTA multimodal baselines to demonstrate its effectiveness:
\emph{1)} UMT~\cite{yu-etal-2020-improving};
\emph{2)}  UMGF~\cite{zhang-UMGF};
\emph{3)} MEGA~\cite{multimodal-re}; 
\emph{4)}  MKGformer.
Apart from previous multimodal approaches, we also compare {\ours} with typical continual learning methods for a fair comparison as follows:
\emph{1)} \textbf{Vanilla} fine-tunes a BERT model on new task data without memory, acting as a \textbf{lower bound} for catastrophic forgetting.
\emph{2)} \textbf{Joint Training} retains all data in memory, retraining the MKGformer for each task, establishing an upper-performance limit.
\emph{3)} \textbf{EWC} constrains critical parameter shifts to preserve performance on prior tasks.
\emph{4)} \textbf{EMR} combines new task data with a memory of key past samples for incremental learning.
\emph{5)} \textbf{EMAR-BERT} employs reconsolidation and activation techniques to address catastrophic forgetting.
\emph{6)} \textbf{RP-CRE} represents the forefront in continual relation extraction, using stored relation samples to refine prototypes.
\emph{7)} \textbf{ExtendNER} applies KD, leveraging an existing NER model to guide the learning of a subsequent model.

\subsection{\textbf{Performance on IMRE Benchmark }} 
\label{sec:results}

Experiments on the IMRE benchmark (Table~\ref{tab:exp-mnre}) yield several insights: \emph{(1)} Fine-tuning unimodal BERT (Vanilla approach) with new examples leads to performance degradation due to overfitting and catastrophic forgetting. Surprisingly, multimodal models, expected to outperform Vanilla, delivered inferior results, emphasizing the need for research in continual multimodal learning.
\emph{(2)} Our method {\ours} outperforms all existing MKGC models. While other continual learning approaches, utilize memory modules and sampling strategies to reduce forgetting, they are outstripped by {\ours} in the 10-split IMRE benchmark, highlighting our method's effective use of multimodal interactions.

\begin{figure}[!]
    \centering
\includegraphics[width=0.35\textwidth]{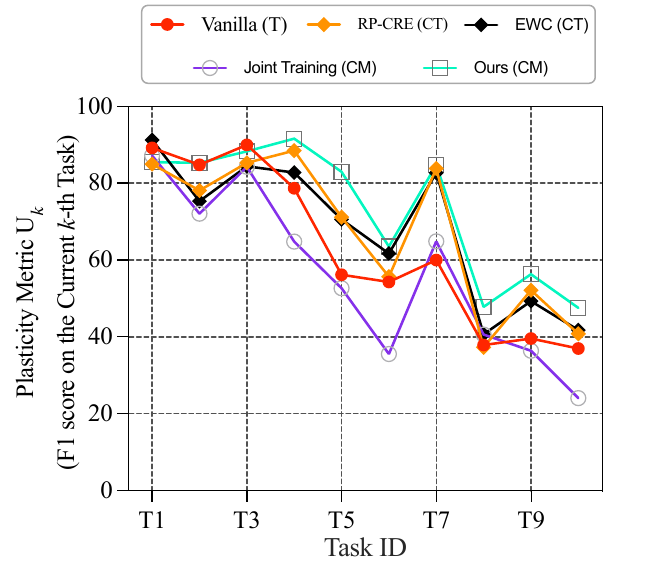}%
    \caption{Performance in plasticity on the IMRE Benchmark.}
    \label{fig:plasticity}
\end{figure}

\subsection{\textbf{Performance on IMNER Benchmark}}

In this section, we thoroughly compare {\ours} with baseline methods across two task orders, detailed in Table~\ref{tab:exp_mner}. The insights are as follows: \textbf{(1) Overall performance:} Despite variations in MKGC model performance, these models generally lag behind unimodal BERT in continual MNER tasks, highlighting unresolved challenges in multimodal continual learning. Yet, {\ours} significantly outshines all competing methods on the IMNER benchmark, demonstrating its robustness and ability to overcome the limitations of previous MKGC approaches in continual settings. \textbf{(2) Task order robustness:} To test {\ours}'s robustness and order independence, we evaluate it on two entity-type permutations: 
``PER $\rightarrow$ ORG $\rightarrow$ LOC $\rightarrow$ MISC'' and ``PER $\rightarrow$ LOC $\rightarrow$ ORG $\rightarrow$ MISC''.
{\ours} consistently tops baselines across permutations, indicating it is not bound to a particular order and can generalize effectively. This across-the-board superiority on the IMNER benchmark confirms the method's effectiveness.
\begin{figure}[!htb]
    \centering
\includegraphics[width=0.35\textwidth]{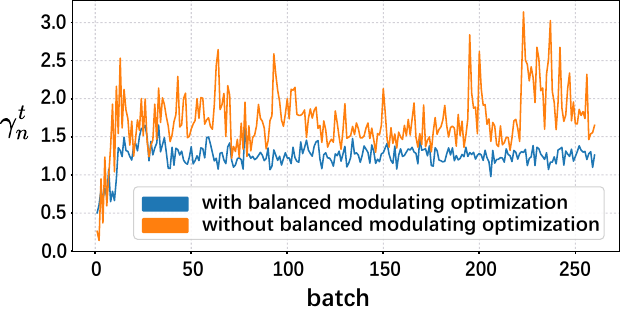}
    \caption{ Change of contribution ratio $\gamma^{t}_{n}$ during training.\label{fig:balance}}
\end{figure}
\textbf{(3)} The ``M-[]'' series methods surpass both RP-CRE and our {\ours} but do not reach the performance levels of SOTA unimodal continual RE methods, suggesting that simple transfer-based strategies are inadequate for optimal performance.

\subsection{Ablation Study and Analysis}

\paragraph{Effect of Each Component.}
Table~\ref{ablation} reveals that each component generally enhances model performance. Specifically, the ``MM'' strategy boosts the average forgetting metric by \textbf{28.5\%}, resonating with evidence that rehearsal is effective for continual KGC. ``GM'' leads to a \textbf{22.2\%} increase in F1 score, suggesting the necessity of balancing learning rates across modalities to reduce forgetting. ``AD'' yields a \textbf{10.4\%} F1 score improvement, indicating that preserving attention patterns aids in retaining prior knowledge. ``MI'' shows a \textbf{5.7\%} F1 score gain, confirming its crucial role in consistent learning.
Notably, omitting ``MI'' resulted in a temporary performance spike on the second task, potentially due to the self-attention mechanism's efficacy in short-term learning, enhanced by attention distillation. However, our findings suggest this approach is less suitable for longer task sequences. The performance declines across all tasks when the other components are removed, further validating the effectiveness of each proposed element.

\paragraph{Plasticity Assessment.}
Our evaluation of model plasticity, depicted in Figure~\ref{fig:plasticity}, indicates that {\ours} surpasses other models employing continual learning strategies like RP-CRE and EWC. We found that Joint Training exhibits the lowest plasticity due to its reliance on replaying all previous tasks' data, which hampers the model's ability to adapt to new tasks. The results highlight the superior plasticity of {\ours}, which outperforms other continual learning approaches and competes with leading multimodal methods. Through attention distillation, {\ours} strikes a balance between maintaining past knowledge and adapting to new information, thereby mitigating catastrophic forgetting effectively.




\paragraph{Imbalance Modulation Analysis.}
Our evaluation investigates our method's capability to mitigate training imbalances by monitoring the discrepancy ratio ${\gamma}{n}^{t}$, which reflects inter-modality disparity. Figure~\ref{fig:balance} demonstrates that our method successfully minimizes ${\gamma}{n}^{t}$, indicating its effectiveness in rectifying the common issue of modality imbalance in datasets. Through nuanced modulation, our approach ensures equitable learning across modalities, promoting a balanced contribution to the learning process.

\begin{figure}[htbp!]
    \centering
\includegraphics[width=0.48\textwidth]{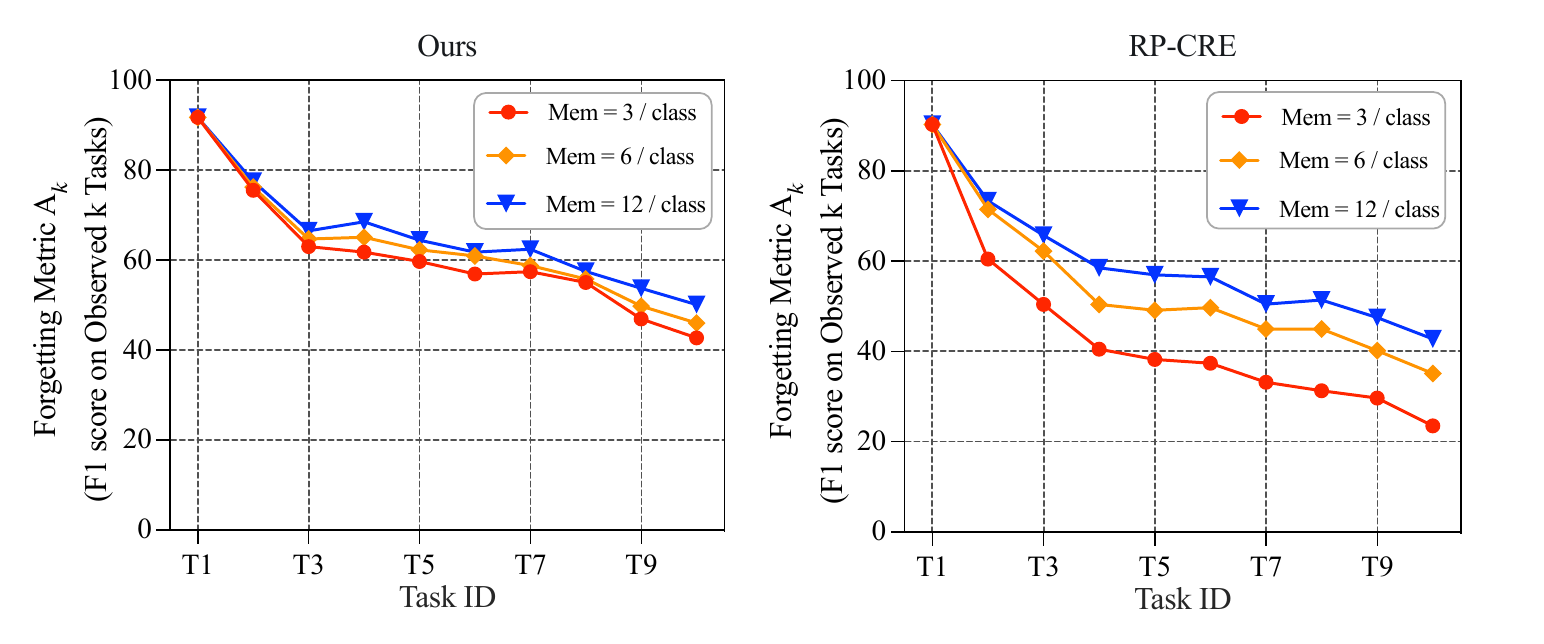}
    \caption{Analysis on rehearsal size.}
    \label{fig:memory}
\end{figure}

\begin{figure}[!htbp]
    \centering
\includegraphics[width=0.35\textwidth]{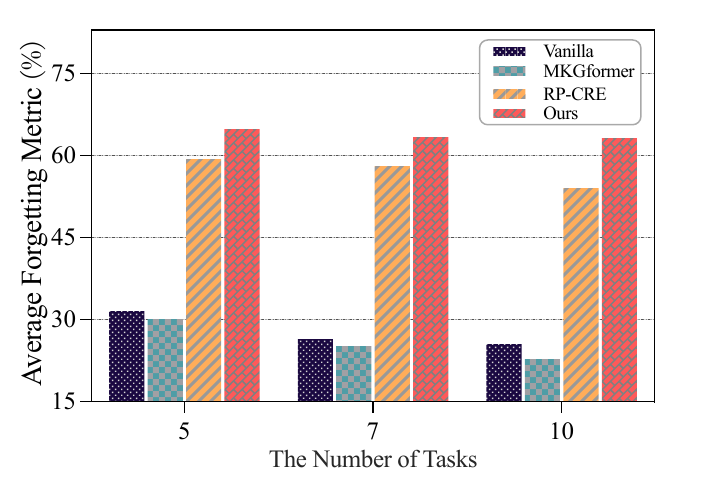}
    \caption{Sensitive Analysis on Task Numbers. \label{fig:tasknumber}}
\end{figure}


\paragraph{Model Dependence on Rehearsal Size.}
The performance of rehearsal-based continual MKGC models is inherently linked to the rehearsal size, which governs the volume of training samples preserved. We assessed our model's robustness by evaluating its performance under varying rehearsal sizes. Our {\ours} model consistently outperforms competing methods on the IMRE benchmark, regardless of the allocated rehearsal size, as depicted in Figure~\ref{fig:memory}. This steadfastness highlights our method's capability to maintain performance even when faced with constraints on rehearsal size. Remarkably, the superiority of our model becomes more apparent with smaller rehearsal sizes, showcasing its effective utilization of limited memory resources.


\paragraph{Sensitive Analysis on Task Numbers.}
We assess how the number of tasks impacts the {\ours} model's performance, using the IMRE benchmark with 5, 7, and 10 tasks. All methods, including RP-CRE, MKGformer, and Vanilla, were tested under uniform experimental conditions: identical random seeds, hyperparameters, and task sequences.
{\ours} demonstrates superiority over RP-CRE and other baselines for all task quantities, showcasing consistent performance regardless of the number of tasks. This consistency confirms the robustness and adaptability of {\ours} for continual MRE.


\section{Conclusion and Future Work}

Our study introduces the novel concept of continual MKGC, addressing the critical and practical challenge of continuously recognizing new entity categories and relations within a knowledge graph. We present a benchmark for MKGC and propose a unique approach named {\ours}, which adeptly combats the dual challenges of catastrophic forgetting and plasticity, central issues in continual learning. {\ours} employs a harmonized multimodal training approach to improve the detection of novel patterns, alongside a synergistic multimodal interaction with attention distillation to effectively retain previous knowledge. Comprehensive experiments and analysis demonstrate the superiority of {\ours} over existing techniques in the context of continual learning. 
Future work will aim to expand our approach to a broader range of MKGC and investigate rehearsal-free strategies for continual MKGC.

\section*{Acknowledgments}

We would like to express gratitude to the anonymous reviewers for kind comments. This work was supported by the National Natural Science Foundation of China (No. 62206246), the Fundamental Research Funds for the Central Universities (226-2023-00138), Zhejiang Provincial Natural Science Foundation of China (No. LGG22F030011), Yongjiang Talent Introduction Programme (2021A-156-G), CCF-Baidu Open Fund, and Information Technology Center and State Key Lab of CAD\&CG, Zhejiang University.

\appendix





\bibliographystyle{named}
\bibliography{ijcai24}

\end{document}


\maketitle

\appendix

\section{Convergence Analysis in Multimodal Learning}
\label{subsec:convergence_analysis}

Thus, we present an analysis of the imbalanced convergence rates in multimodal models.
Representatively, we take MRE sub-task as example\footnote{Here MNER sub-task also can be regarded as the token-level classification.} and denote training dataset $\mathcal{D}_k=\{x_{i}, y_{i}\}_{i=1,2...N}$. Each $x_{i}$ contains two inputs from visual and textual modalities as $x_{i}=(x^{v}_{i}, x^{t}_{i})$, respectively. $y_i \in \left\{1,2,\cdots,|\mathcal{R}_{\Delta k}|\right\}$, where $|\mathcal{R}_{\Delta k}|$ denotes the total number of relation types at the $k$-th task that the model has to evaluate. We use two encoders $f^{v}(\theta^{v},\cdot)$ and $f^{t}(\theta^{t},\cdot)$ to extract visual and textual features, where $\theta^{v}$ and $\theta^{t}$ are the parameters of encoders. 
Since many MKGC models utilize concatenation as a vanilla fusion method, we let $W \in \mathbb{R}^{|\mathcal{R}_{\Delta k}| \times (d_{v}+d_{t})}$  denote the parameter of the final linear classifier, where $d_v$ and $d_t$ is the embedding dimension of image and text respectively. Then we define the output logit of the MRE model as follows:
\begin{equation}
\label{output}
\footnotesize
\mathcal{M}(x_{i})=W[f^{v}(\theta^{v},x^{t}_{i});f^{t}(\theta^{t},x^{t}_{i})].
\end{equation}
We further divide $W$ as the combination of two blocks: $[W^{v}, W^{t}]$ to individually analyze each modality's optimization process, thus, rewrite the Equation~\ref{output} as:
\begin{equation}
\label{output_re}
\footnotesize
\mathcal{M}(x_{i})=W^{v}\cdot f^{v}(\theta^{v},x^{v}_{i})+ W^{t}\cdot f^{t}(\theta^{t},x^{t}_{i}).
\end{equation}
The cross-entropy loss of the classification model  is represented as  $\mathcal{L}_{CE}=- \frac{1}{N} \sum^{N}_{i=1}  \text{log} \frac{e^{f(x_{i})_{y_i}}}{\sum^C_{j=1}e^{f(x_{i})_{j}}}$. 
We update $W^{v}$ and the parameters of encoder $f^{v}(\theta^{v},\cdot)$ with gradient descent optimization method (similarly for $W^{t}$ and $f^{t}(\theta^{t},\cdot)$ ) as follows:

\begin{equation}
\label{update_W}
\footnotesize
\begin{aligned}
W_{n+1}^{v} &=W_{n}^{v}-\eta\nabla_{W^{v}} \mathcal{L}_{CE}(W_{n}^{v})\\
&=W_{n}^{v}-\eta \frac{1}{N} \sum^{N}_{i=1} \frac{\partial \mathcal{L}_{CE} }{\partial \mathcal{M}(x_{i})} f^{v}(\theta^{t},x^{v}_{i}),
\end{aligned}
\end{equation}

\begin{equation}
\label{update_f}
\footnotesize
\begin{aligned}
\theta_{n+1}^{v} &=\theta_{n}^{v}-\eta\nabla_{\theta^{v}} L(\theta_{n}^{v})\\
&=\theta_{n}^{v}-\eta  \frac{1}{N} \sum^{N}_{i=1} \frac{\partial \mathcal{L}_{CE} }{\partial \mathcal{M}(x_{i})}  \frac{\partial( W^{v}_{n}\cdot f^{v}_{n}(\theta^{v},x^{}_{i}))}{\partial \theta_{n}^{v}},
\end{aligned}
\end{equation}
where $\eta$ denotes the learning rate. We can see from Equations~\ref{update_W} and~\ref{update_f} that the optimization of $W^{v}$ and $f^{v}$ is almost independent of that of the other modality, except for the term regarding the training loss ($\frac{\partial L_{CE} }{\partial \mathcal{M}(x_{i})}$), leading to different levels of convergence. 
Furthermore, in the context of continual learning, this characteristic becomes more pronounced, resulting in inconsistent forgetting of past tasks and imbalanced learning dynamics for the current task among multiple modalities. This aspect aligns with the MKGC model's ability to mitigate catastrophic forgetting and effectively learn from examples with new entity categories and relations.

\section{Algorithm Flow of {\ours} Training}

\begin{equation}
\label{loss:distill}
\begin{aligned}
\mathcal{L}_{\text{AD}} &= \mathcal{L}_{\text{AD-width }}^v +\mathcal{L}_{\text{AD-width}}^t 
\end{aligned}
\end{equation}

\begin{algorithm}[h!]
\caption{ Training process for the $k$-th task}
\small
\begin{algorithmic}[1] 
    \Require the training set $D_k$, the current memory $ {\mathcal{O}}$, the model $\theta$, and the last model $\theta^{(k-1)}$.
    \For{$epoch=1, \ldots, epochs_{c}$} \Comment{\blue{Train current task}}
        \For {$batch=1, \ldots , batches_{c}$} 
            \State \textsc{Sample} $batch_{current}$ from $D_k$ 
            \If{$k=1$}
                \State \textsc{Input} $batch_{current}$ into model $\theta^{(k)}$
            \Else
                \State \textsc{Input} $batch_{current}$ into model $\theta^{(k-1)}$ and $\theta^{(k)}$ 
                \State \textsc{Store} attention maps from model $\theta^{(k-1)}$ and $\theta^{(k)}$ 
                \State \textsc{Using} Equation~(\ref{loss:distill}) to calculate attention distillation loss
            \EndIf
            \State \textsc{Calculate} the gradient of $\theta^{(k)}$
            \State \textsc{Do} modality balance according to \S Section-4.2 and \textsc{Update} $\theta^{(k)}$
        \EndFor
    \EndFor
    \State \textsc{Select} samples from $D_k$ and store them in $\hat{o}$ 

    \State $ {\mathcal{O}} \leftarrow \hat{o} \cup  {\mathcal{O}}$ \Comment{\blue{Update memory}}
    \For{$epoch=1, \ldots, epochs_{m}$} \Comment{\blue{Train in memory}}
        \For {$batch=1, \ldots , batches_{m}$} 
            \State \textsc{Sample} $batch_{memory}$ from $ {\mathcal{O}}$ and \textsc{Input} it into model $\theta^{(k)}$
            \State \textsc{Calculate} the gradient of $\theta^{(k)}$
            \State \textsc{Do} modality balance according to \S Section-4.2 and \textsc{Update} $\theta^{(k)}$
        \EndFor
    \EndFor 
\end{algorithmic}
\label{alg:Framwork} 
\normalsize
\end{algorithm}
